# Uncertainty of Joint Neural Contextual Bandit

**Hongbo Guo**
hongbog@meta.com
Meta AI

**Zheqing Zhu**
billzhu@meta.com
Meta AI

## Abstract

Contextual bandit learning is increasingly favored in modern large-scale recommendation systems. To better utlize the contextual information and available user or item features, the integration of neural networks have been introduced to enhance contextual bandit learning and has triggered significant interest from both academia and industry. However, a major challenge arises when implementing a disjoint neural contextual bandit solution in large-scale recommendation systems, where each item or user may correspond to a separate bandit arm. The huge number of items to recommend poses a significant hurdle for real world production deployment. This paper focuses on a joint neural contextual bandit solution which serves all recommending items in one single model. The output consists of a predicted reward $\mu$, an uncertainty $\sigma$ and a hyper-parameter $\alpha$ which balances exploitation and exploration, e.g., $\mu + \alpha\sigma$.

The tuning of the parameter $\alpha$ is typically heuristic and complex in practice due to its stochastic nature. To address this challenge, we provide both theoretical analysis and experimental findings regarding the uncertainty $\sigma$ of the joint neural contextual bandit model. Our analysis reveals that $\alpha$ demonstrates an approximate square root relationship with the size of the last hidden layer $F$ and inverse square root relationship with the amount of training data $N$, i.e., $\sigma \propto \sqrt{\frac{F}{N}}$. The experiments, conducted with real industrial data, align with the theoretical analysis, help understanding model behaviors and assist the hyper-parameter tuning during both offline training and online deployment.

## 1 Introduction

The Bandit algorithms (Lattimore & Szepesvári, 2020) as part of the reinforcement learning algorithms family (Sutton & Barto, 1998) have emerged as powerful tools and favored approaches in the realm of large-scale recommendation systems (Elena et al., 2021). A plethora of bandit learning algorithms has emerged, including various methodologies such as non-contextual upper confidence bound (UCB), linear contextual upper confidence bound, neural network-based bandits, and epistemic neural recommendation algorithms (Li et al., 2010; Zhou et al., 2020; Xu et al., 2020; Zhu & Van Roy, 2023; Guo et al., 2023; Nguyen-Thanh et al., 2019; Huang et al., 2022; Zhu et al., 2023; Silva et al., 2022; Zheng et al., 2018). The bandit solution balances exploration and exploitation by iteratively selecting actions based on estimated rewards and uncertainties. Initial parameters are set, and for each round, actions are chosen based on the corresponding reward's posterior distribution. This way, rewards are observed and model parameters are updated iteratively. Extensive analysis had been conducted mostly on regret of these bandit algorithms (Beygelzimer et al., 2017; Bubeck et al., 2012; Lu et al., 2020). Some examples of early adoptions of bandit solutions in video recommendations are described in Covington et al. (2016); Gomez-Uribe & Hunt (2015), and then some further developments as in Guo et al. (2020); Zhang et al. (2022).

Among those literatures, the pioneer work of linear bandit algorithm described in Li et al. (2010) assumes a linear relationship between contextual features and the expected rewards. During this process, a linear matrix need to be maintained iteratively and its inverse need to be calculated which is computation heavy (Tveit, 2003). Based on this pioneer work, to further utilize the power of deep learning and to introduce non-linearity, some works such as Xu et al. (2020); Zhou et al. (2020) extended linear bandit with a neural network to pre-process the raw features before they are fed to the linear bandit module. The raw features are transformed by an multi layer perception (MLP) module and the output of the MLP's last hidden layer are fed to a classical linear contextual module.

In some large-scale recommendation systems, the sheer volume of items to be recommended, often numbering in the millions, presents a considerable challenge. Consequently, training and deploying a disjoint neural contextual bandit solution, where each item may be treated as a bandit arm[1], requests the maintenance of millions or even billions of matrices and as a result may prove impractical. An alternative approach, where there is no need for training a deep learning model, entails utilizing a non-contextual UCB algorithm (Nguyen-Thanh et al., 2019; Guo et al., 2023) with feature clusters. In such approaches, users or items can be clustered by available features, and then each cluster is treated as a bandit arm. Such non-neural network approaches still come with some drawbacks, e.g., the features may not be most efficiently utilized, and clustering may not be updated timely.

On the other hand, a joint deep neural network bandit is able to both utilize features efficiently and avoid maintaining an individual matrix for each recommending item. In such solution, the score for all items follows the same format: a predicted reward $\mu$, an associated uncertainty $\sigma$ and a critical hyper-parameter $\alpha$ which governs the balance between exploration and exploitation. From this single model, the score $\mu + \alpha\sigma$ can be generated for each item. The uncertainty term $\sigma$ is inherently stochastic and fluctuates during both training and deployment phases. This poses challenges in real industrial deployment for determining the optimal value of $\alpha$, because $\alpha$ balances $\mu$ with $\sigma$ where the value range of the former may be prior known approximately[2] but the latter may not. Usually in production the tuning of $\alpha$ often relies on heuristic methods and is lack of theoretical justification. This paper aims to address this gap. By delving into the theoretical investigations and empirical observations, this study endeavors to enhance the understanding of how a joint neural contextual bandit model operates, and offer valuable insights into the effective tuning of the hyper-parameter.

The remainder of the paper is organized as follows: Section 2 describes the joint neural contextual bandit solution, analyzing feature distribution and uncertainty fluctuation. Section 3 presents the experimental results obtained from industrial data. Finally, Section 4 concludes this work.

## 2 Joint Neural Contextual Bandit and Uncertainty

### 2.1 Joint neural contextual bandit

A joint neural contextul bandit model architecture consists of two modules: a multi-layer perception (MLP) or neural network module which consumes the raw features and outputs processed features **f**, and a linear bandit module which consumes **f** and outputs predicted reward $\mu$ and uncertainty $\sigma$. This architecture is depicted in Fig. 1. All items to be recommended share the same model, i.e., only one MLP module need to be trained, and only one matrix inverse in the linear bandit module need to be maintained. The MLP module with the trainable bottom layers can be updated through gradient descent and back-propagation by leveraging the optimizers of modern deep learning frameworks such as PyTorch, TensorFlow or MXNet(Paszke et al., 2019; Chen et al., 2015; Pang et al., 2020). In contrast, the linear module constituting the final layer, is not updated through gradient descent but via matrix computations and matrix inverse operations. These matrix operations can be accelerated by modern matrix algorithms (Courrieu, 2008) but are still with considerable computation cost. Fortunately, in a joint neural contextual bandit model we maintain only one single such matrix for all recommending items.

Refer to Fig. 1, officially each training data point contains a raw feature vector and a scalar reward. The raw feature vector goes through MLP module which outputs $\mathbf{f} \in \mathbb{R}^{F \times 1}$ as a processed feature vector matching the size of the last hidden layer of the MLP module. Name $\mathbf{F} \in \mathbb{R}^{F \times N}$ as a stack of $N$ feature vectors from $N$ training data points, then we have a linear fitting problem

$$\mathbf{y} = \mathbf{F}^T \mathbf{w} + \mathbf{e} \qquad (1)$$

where $\mathbf{y} \in \mathbb{R}^{N \times 1}$ is the ground truth reward, $\mathbf{F}^T \in \mathbb{R}^{N \times F}$ is the stack of MLP processed features, $\mathbf{w} \in \mathbb{R}^{F \times 1}$ is the linear weights to be derived, and $\mathbf{e} \in \mathbb{R}^{N \times 1}$ is fitting error. By finding the optimal $\mathbf{w}$, we can derive the joint contextual bandit algorithm as

$$\begin{cases} \mu_{\mathbf{w}} = \left(\mathbf{F}\mathbf{F}^T + \gamma \mathbf{I}\right)^{-1} \mathbf{F}\mathbf{y} \\ \Sigma_{\mathbf{w}} = \left(\mathbf{F}\mathbf{F}^T + \gamma \mathbf{I}\right)^{-1} \end{cases} \qquad (2)$$

---

[1]Depending on the scenario, item or user, or even <user, item> pair maybe considered as a bandit arm.
[2]In our application, reward is conversion rate which is prior known to be mostly within a specific range, e.g., 0 0.1.

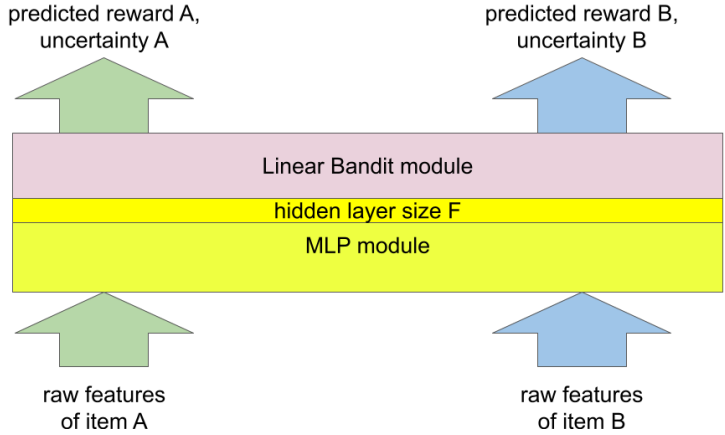

Figure 1: Joint neural contextual bandit.

Here $\mu_{\mathbf{w}} \in \mathbb{R}^{F \times 1}$ and $\Sigma_{\mathbf{w}} \in \mathbb{R}^{F \times F}$. The scaling factor $\gamma \mathbf{I} \in \mathbb{R}^{F \times F}$ is related to the Gaussian assumption on prior distribution of weights, Prob($\mathbf{w}$). As regularization item it can help reduce overfitting. Also, it avoids numerically instability in case of inverting a low rank matrix [3]. More details to derive Eq. 2 refer to Appendix A and B.

On an inference data, we get its predicted reward and uncertainty through its feature vector $\mathbf{f}$ by

$$\begin{cases} \mu = \mathbf{f}^T \mu_{\mathbf{w}} = \mathbf{f}^T \left(\mathbf{F}\mathbf{F}^T + \gamma \mathbf{I}\right)^{-1} \mathbf{F}\mathbf{y} \\ \sigma^2 = \mathbf{f}^T \Sigma_{\mathbf{w}} \mathbf{f} = \mathbf{f}^T \left(\mathbf{F}\mathbf{F}^T + \gamma \mathbf{I}\right)^{-1} \mathbf{f} \end{cases} \qquad (3)$$

Eventually the uncertainty is derived as $\sigma = \sqrt{\mathbf{f}^T \left(\mathbf{F}\mathbf{F}^T + \gamma \mathbf{I}\right)^{-1} \mathbf{f}}$ and it can be used for exploration together with hyper-parameter $\alpha$ to form $\mu + \alpha \sigma$.

### 2.2 Features distribution and approximation

For a data point, the MLP output $\mathbf{f}$ consists of two components, a centroid $\mathbf{x}$ and a variant $\mathbf{n}$:

$$\mathbf{f} = \mathbf{x} + \mathbf{n} \qquad (4)$$

where $\mathbf{x}$ is common to all data and normalized

$$\mathbf{x}^T \mathbf{x} = \|\mathbf{x}\|^2 = 1 \qquad (5)$$

---

[3]In real industrial scenario, this identity matrix is not critical because features are not identical and as a result the matrix will not be low rank. Further, in case of inverting a low rank matrix, the Moore-Penrose inverse or `torch.linalg.pinv` can be applied to avoid numerical stability issue.

while $\mathbf{n}$ is a vector of independent and identically distributed (i.i.d.) random variables from normal distribution with zero mean. More officially, $\mathbf{n} = (n_1, \ldots, n_f, \ldots, n_F)$ where each $n_f$ follows

$$n_f \sim \text{i.i.d. } \mathcal{N}(0, \sigma_n^2) \tag{6}$$

This assumption is based on several observations. Different items can share some commonness due to the same data collection policy. Further, $\mathbf{f}$ is the output of the last hidden layer of well trained MLP module. Due to the central limit theorem, assuming the network has a large number of neurons and the inputs are sufficiently complex, then the combination of non-linear transformations and homogeneous training leads to the outputs of the last hidden layer of a well-trained MLP approximately following i.i.d. normal distribution.

With the features of multiple data points $\mathbf{F} = [\mathbf{f}_1, \cdots, \mathbf{f}_N] \in \mathcal{R}^{F \times N}$ where $\mathbf{F} = \mathbf{X} + \mathbf{N}$ and $\mathbf{X} = [\mathbf{x}, \cdots, \mathbf{x}] \in \mathcal{R}^{F \times N}$, $\mathbf{N} = [\mathbf{n}_1, \cdots, \mathbf{n}_N] \in \mathcal{R}^{F \times N}$, we have

$$\mathbf{F}\mathbf{F}^T = (\mathbf{X} + \mathbf{N})(\mathbf{X} + \mathbf{N})^T = (\mathbf{X} + \mathbf{N})(\mathbf{X} + \mathbf{N})^T = \mathbf{X}\mathbf{X}^T + \mathbf{X}\mathbf{N}^T + \mathbf{N}\mathbf{X}^T + \mathbf{N}\mathbf{N}^T \tag{7}$$

That is, $\mathbf{N}$ is the delta to the feature centroids $\mathbf{X}$. When it is small, we have approximately $\mathbf{F} \approx \mathbf{X}$. Note $\mathbf{X}$ has identical columns as all data points share the same centroid. $\mathbf{N}$ does not have identical columns. Statistically, the expectation of $\mathbf{A}$ is

$$\mathbb{E}\left(\mathbf{F}\mathbf{F}^T\right) = \mathbb{E}\left(\mathbf{X}\mathbf{X}^T + \mathbf{X}\mathbf{N}^T + \mathbf{N}\mathbf{X}^T + \mathbf{N}\mathbf{N}^T\right) = \mathbb{E}\left(\mathbf{X}\mathbf{X}^T\right) + \mathbf{I}\Sigma_n \tag{8}$$

Without loss of generality and for simplicity, our analysis assumes the variance in each MLP output dimension is uniform, $\Sigma_n = \text{Diag}[\sigma_1 \cdots \sigma_F]$ where $\sigma_1 = \cdots = \sigma_F = \epsilon$. This reduces Eq. 8 into

$$\mathbb{E}\left(\mathbf{F}\mathbf{F}^T\right) = \mathbb{E}\left(\mathbf{X}\mathbf{X}^T\right) + \epsilon \mathbf{I} \tag{9}$$

Usually there is also a regularization term $\gamma \mathbf{I}$ which comes from the prior assumption of $\mathbf{w}$ as in Eq. 2 and 3, but it can be mathematically merged into $\epsilon \mathbf{I}$ as one single term for simplicity. From here, we name

$$\mathbf{A} = \mathbf{X}\mathbf{X}^T \tag{10}$$

and focus on the analysis of $\mathbf{X}\mathbf{X}^T + \epsilon \mathbf{I}$ (or $\mathbf{X}\mathbf{X}^T + \gamma \mathbf{I}$, indistinguishable between these two) without losing generality.

### 2.3 Uncertainty of the special case : inference data with $\mathbf{f} = \mathbf{x}$

In a special case when the feature vector is identical, i.e., the variance of feature $\mathbf{f}$ being zero, we have $\mathbf{f} = \mathbf{x}$, $\mathbf{n} = 0$ and

$$(\mathbf{A} + \epsilon \mathbf{I})\mathbf{x} = \left(\mathbf{X}\mathbf{X}^T + \epsilon \mathbf{I}\right)\mathbf{x} = \left(N\mathbf{x}\mathbf{x}^T + \epsilon \mathbf{I}\right)\mathbf{x} = N\mathbf{x}\mathbf{x}^T\mathbf{x} + \epsilon \mathbf{I}\mathbf{x} = N\mathbf{x} + \epsilon\mathbf{x} = (N + \epsilon)\mathbf{x} \tag{11}$$

where $N$ is the number of training data points. Eq. 11 means

$$\lambda = N + \epsilon \tag{12}$$

where $\lambda$ is the eigenvalue of $\mathbf{A} + \epsilon \mathbf{I}$ and its corresponding eigenvector is $\mathbf{x}$. When matrix inverse exists, the eigenvalues of the inverse matrix are equal to the inverse of the eigenvalues of the original matrix[4], we get

$$(\mathbf{A} + \epsilon \mathbf{I})^{-1}\mathbf{x} = \frac{1}{N + \epsilon}\mathbf{x} \tag{13}$$

---
[4] In our experiments, features of different items are different, and the regularization item $\epsilon$ inside $\mathbf{X}\mathbf{X}^T + \epsilon \mathbf{I}$ is set to non-zero. Thus the inverse $(\mathbf{A} + \epsilon \mathbf{I})^{-1}$ always exist.

and then straightforwardly

$$\mathbf{x}^T(\mathbf{A}+\epsilon\mathbf{I})^{-1}\mathbf{x} = \mathbf{x}^T \frac{1}{N+\epsilon}\mathbf{x} = \frac{1}{N+\epsilon}\mathbf{x}^T\mathbf{x} = \frac{1}{N+\epsilon}\|\mathbf{x}\|^2 = \frac{1}{N+\epsilon} \quad (14)$$

$$\sigma = \frac{1}{\sqrt{\mathbf{x}^T(\mathbf{A}+\epsilon\mathbf{I})^{-1}\mathbf{x}}} = \frac{1}{\sqrt{N+\epsilon}} \quad (15)$$

When our training data amount $N$ is huge such as thousands or even millions Eq. 15 approximately reduces to $\sigma \approx \frac{1}{\sqrt{N}}$. That is, uncertainty linearly decreases with the increase of square root of training data amount[5].

### 2.4 Uncertainty of the general case: inference data with $\mathbf{f} \neq \mathbf{x}$

A more general case is that the feature vector consists of both centroid and variation, i.e., $\mathbf{f} = \mathbf{x} + \mathbf{n}$. In this case we have

$$\begin{aligned}\sigma^2 &= \mathbf{f}^T(\mathbf{A}+\epsilon\mathbf{I})^{-1}\mathbf{f} \\ &= (\mathbf{x}+\mathbf{n})^T(\mathbf{A}+\epsilon\mathbf{I})^{-1}(\mathbf{x}+\mathbf{n}) \\ &= \mathbf{x}^T(\mathbf{A}+\epsilon\mathbf{I})^{-1}\mathbf{x} + \mathbf{x}^T(\mathbf{A}+\epsilon\mathbf{I})^{-1}\mathbf{n} + \mathbf{n}^T(\mathbf{A}+\epsilon\mathbf{I})^{-1}\mathbf{x} + \mathbf{n}^T(\mathbf{A}+\epsilon\mathbf{I})^{-1}\mathbf{n}\end{aligned} \quad (16)$$

Due to $\mathbb{E}(\mathbf{n}) = 0$, we have[6]

$$\mathbb{E}\left(\mathbf{x}^T(\mathbf{A}+\epsilon\mathbf{I})^{-1}\mathbf{n}\right) = \mathbb{E}\left(\mathbf{n}^T(\mathbf{A}+\epsilon\mathbf{I})^{-1}\mathbf{x}\right) = 0 \quad (17)$$

and then approximately Eq. 16 reduces to [7]

$$\sigma^2 \approx \mathbf{x}^T(\mathbf{A}+\epsilon\mathbf{I})^{-1}\mathbf{x} + \mathbf{n}^T(\mathbf{A}+\epsilon\mathbf{I})^{-1}\mathbf{n} = \frac{1}{N+\epsilon} + \frac{\|\tilde{\mathbf{n}}\|^2}{N+\epsilon} = \frac{1+\sigma_{\tilde{n}}}{N+\epsilon} \quad (18)$$

Here the term $\mathbf{n}^T(\mathbf{A}+\epsilon\mathbf{I})^{-1}\mathbf{n}$ is represented by $\|\tilde{\mathbf{n}}\|^2$. $\mathbf{A}$ is constant because of $\mathbf{X}$ being the centroid, and $\mathbf{n}$ follows a stable distribution, thus $\sigma_{\tilde{n}}$ is stable and can be treated as a constant. When the amount of training data is huge, we can ingore $\epsilon$ and approximately get

$$\sigma \propto N^{-\frac{1}{2}} \quad (19)$$

Further, the argument vector $\mathbf{f}$ has shape $F$, and as a result its corresponding quadratic form $\mathbf{f}^T(\mathbf{A}+\epsilon\mathbf{I})^{-1}\mathbf{f}$ in nature is proportional to the magnitude $F$. That is, the $\sigma^2$ is approximately proportional to the MLP output layer size:

$$\sigma \propto \sqrt{F} \quad (20)$$

More details can be found in Appendix C. Eventually we have

$$\sigma \propto \sqrt{\frac{F}{N}} \quad (21)$$

where $F$ is the MLP output layer size and $N$ is the training data amount[8].

Due to the fact that the MLP as a deep learning module not being transparent, we do not have variance information of the distribution of the MLP output. In practice, we can introduce a heuristic scaling hyper-parameter $C$ and get $\sigma \propto C\sqrt{F/N}$. The heuristic value of $C$ may depend on specific use case, its training dataset and feature variance.

---

[5]Intuitively, this is because feature vectors across huge amount of data points are correlated since they share the same centroid component, but the variation component is random across multiple data points.

[6]Particularly, with a larger $F$ the MLP last layer size, $\mathbf{x}^T(\mathbf{A}+\epsilon\mathbf{I})^{-1}\mathbf{n}$ or $\mathbf{n}^T(\mathbf{A}+\epsilon\mathbf{I})^{-1}\mathbf{x}$ approaches to $0^+$.

[7]We always have $\mathbf{n}^T(\mathbf{A}+\epsilon\mathbf{I})^{-1}\mathbf{n} > 0$ because $(\mathbf{A}+\epsilon\mathbf{I})^{-1}$ is Hermitian and $\mathbf{n} \neq 0$, as a result the quadratic form $\mathbf{n}^T(\mathbf{A}+\epsilon\mathbf{I})^{-1}\mathbf{n} > 0$. Shortly, $\mathbf{n}$ can be decomposed into two components within Range($\mathbf{x}$) and its orthogonal space Range($\mathbf{x}$)$^\perp$ respectively, then the component in Range($\mathbf{x}$) counts for $\|\tilde{\mathbf{n}}\|^2$. For brevity, more details are not included.

[8]Note we cannot directly reuse the logic in Eq. 11 $\sim$ Eq. 15 to conclude that $\sigma$ is only related to $N$ but unrelated to $F$, because $\mathbf{F}$ does not have identical columns (while $\mathbf{X}$ does). In the analysis of $\sigma^2 = \mathbf{f}^T(\mathbf{A}+\epsilon\mathbf{I})^{-1}\mathbf{f}$ it is the variation component $\mathbf{n}$ that mostly contributes to $\sigma \propto \sqrt{F}$.

## 3 Experiment Set Up and Results

In our experiment, we adopted a neural contextual bandit model as in Fig 1 and the output consists of a predicted reward $\mu$ and an uncertainty $\sigma$. The predicted reward is predicted average conversion rate (CVR). Since CVR is usually very small, our $\mu$ is mostly in range of 0.0 to 0.1. In our experiment, we adopted various amount of training data $N$ and various MLP output layer sizes $F$.

In Fig 2, we set a very small $\alpha = 0.01$ with little exploration where the model's output $\mu + \alpha\sigma$ is dominated by exploitation. The output is mostly in the range of 0 to 0.1, basically same as the range of $\mu$, because of this small $\alpha$. On the contrary, we emphasize exploration in Fig 3 where a very large value of $\alpha = 100$ is adopted. The output is dominated by uncertainty $\sigma$ over the predicted reward $\mu$. Overall, Figures 2 and 3 together show that $\alpha$ can hugely affect the model's final output. In order to best configure $\alpha$, we need to know the approximate value range of $\mu$ and $\sigma$. The former term, $\mu$, is relatively stable during both training and deployment because it is predicted CVR and we prior know its rough range. Thus, the anticipation of uncertainty term $\sigma$ is of critical importance to tune the model's hyper-parameter $\alpha$ so that we can reasonably balance up exploration with exploitation.

Fig. 4 shows uncertainty $\sigma$ with different MLP output size $F = 32, 64, 128, 256, 512, 1024$. It is seen that the uncertainty $\sigma$ increases proportionally to the square root of $F$, approximately following $\sigma \propto \sqrt{F}$. This result matches analysis and provides some guidance for us to pre-hand decide the size of last hidden layer. Fig. 5 shows the uncertainty monotonically decreases as the square root of $N$ increases, approximately following $\sigma \propto \frac{1}{\sqrt{N}}$. This is very useful in direction of our planning of offline training and online test. In real application, we first train a model with offline logged data, and then we put model to online serving wherein the model is updated with new data daily. It is the online stage that we need a reasonable *anticipation of the future* value of uncertainty $\sigma$, so as to instruct the configuration of a long term $\alpha$. Further, in some industry scenarios, due to serving and infrastructure limitations, it is not preferred to frequently change model hyper-parameters such as $\alpha$. Instead, it is preferred to set a semi-permanent $\alpha$ without changing it too often. Fig 5 shows the uncertainty drops very quickly at the very beginning with a limited amount of training data, and then it decreases very slowly. This encourage us to adopt some aggressive $\alpha$ during offline training stage, and then we can deploy the trained model to online serving without having to make change to $\alpha$ as the serving infrastructure does not favor such change.

In our experiments, we first train with offline data and figure out an heuristic value of the hyper-parameter $C$ as in $\sigma = \sqrt{\frac{F}{N}}$, then we can use this heuristic value to estimate future $\sigma$. Note here we focus on $\sigma \propto \frac{1}{\sqrt{N}}$ rather than $\sigma \propto \sqrt{F}$ because MLP last layer size $F$ will never be changed once it is set up. On the other hand, $N$ increases day by day. For example, we already trained the model with 200,000 offline data, and then during online serving stage we will have 20,000 new training data daily. We can anticipate that after 5 days of serving the uncertainty will drop from about 0.09 to 0.07 in Fig 5. Further, we can anticipate after two months the uncertainty will drop to 0.004, and after a year drops to 0.001. This indicates that we do not need change $\alpha$ very often, e.g., maybe change by the end of the first two months of serving, and then by the end of the first year.

## 4 Conclusion

In this work, we depicted a joint neural bandit solution whose output consists of a predicted reward term and an uncertainty term, and then presented the challenges of its deployment due to the unknown fluctuation of the uncertainty term. To fill in this gap, we provided theoretical analysis of uncertainty, together with experimental results conducted from real-world industrial data. Our experimental observations align with theoretical analysis, demonstrating that the uncertainty monotonic decreases proportional to the square root of historical data amount, and monotonic increases proportional to the square root of the final hidden layer size. These results serve to guide the development and deployment of joint neural contextual bandit solution in large-scale recommendation systems, and helps to instruct the tuning of key hyper-parameter to balance up exploitation with exploration.

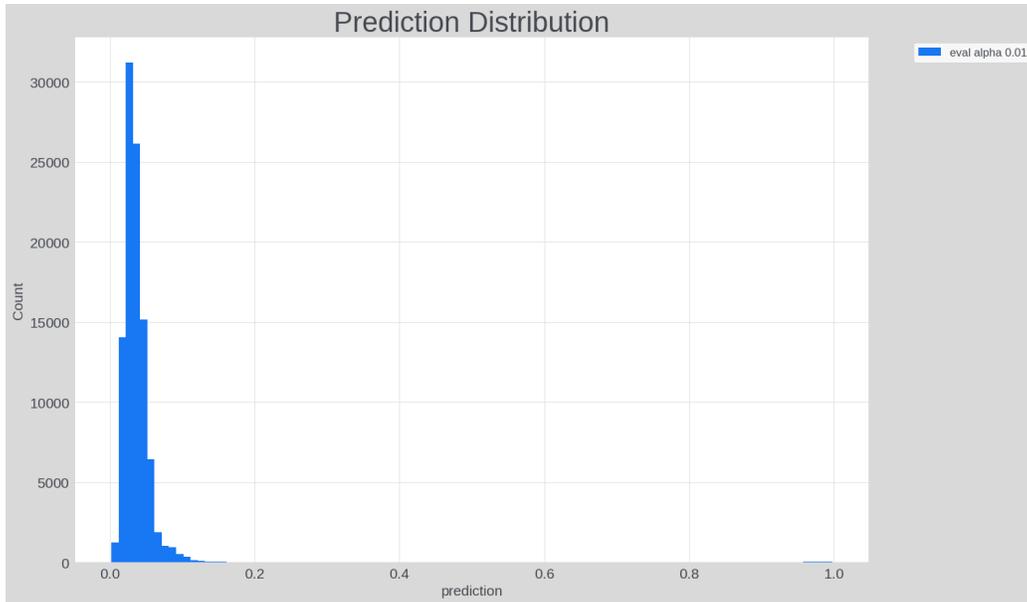

Figure 2: Model's output prediction $\mu + \alpha\sigma$ with $\alpha = 0.01$.

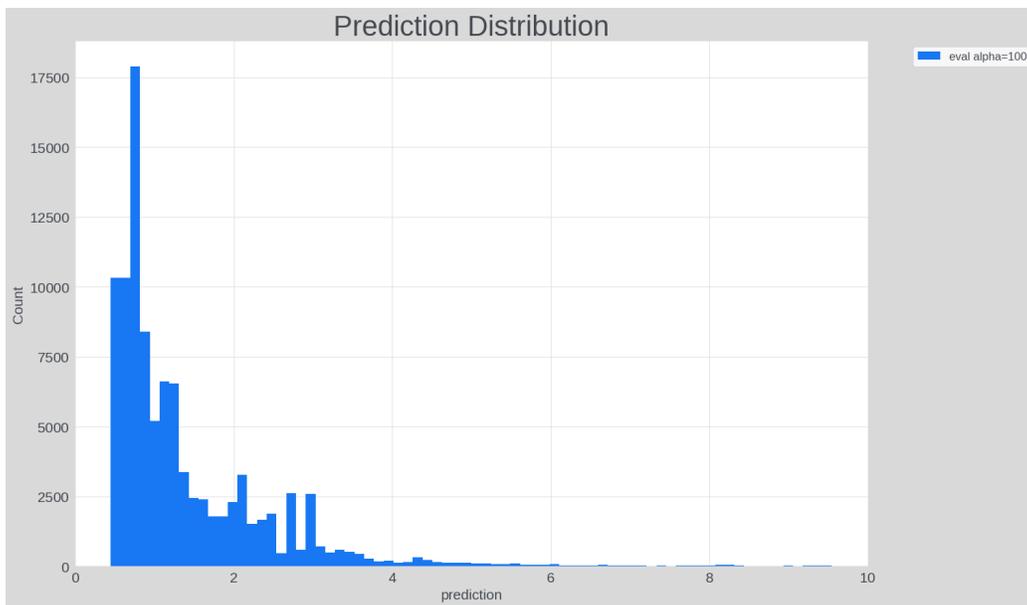

Figure 3: Model's output prediction $\mu + \alpha\sigma$ with $\alpha = 100$.

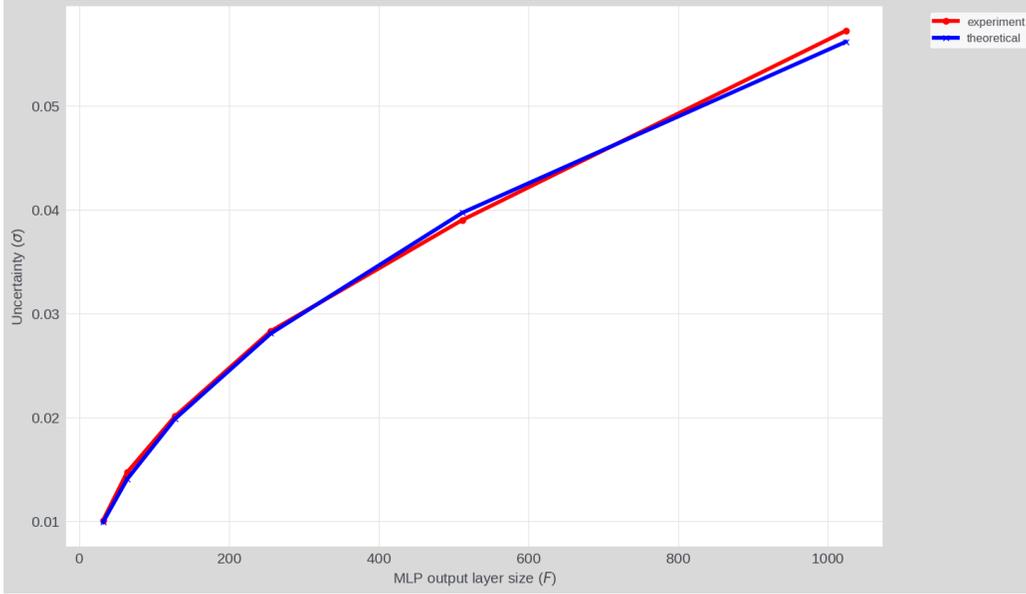

Figure 4: Uncertainty $\sigma$ with the last hidden layer size $F$, $\sigma \propto \sqrt{F}$.

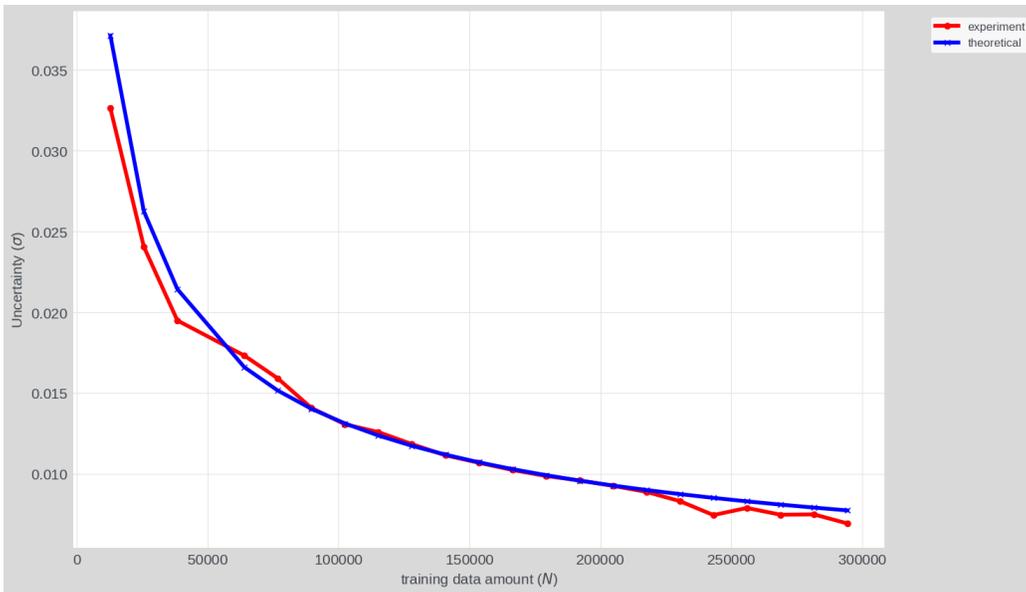

Figure 5: Uncertainty $\sigma$ with training data amount $N$, $\sigma \propto \frac{1}{\sqrt{N}}$.

## A Derivation of $\mu_\mathbf{w}$

By optimizing the posterior distribution

$$\arg\max_\mathbf{w} J(\mathbf{w}) = \arg\max_\mathbf{w} \underbrace{\text{Prob}\left(\mathbf{w}|\mathbf{F},\mathbf{y}\right)}_{\text{posterior}} \propto \underbrace{\text{Prob}\left(\mathbf{y}|\mathbf{F},\mathbf{w}\right)}_{\text{likelihood}} \cdot \underbrace{\text{Prob}(\mathbf{w})}_{\text{prior}} \tag{22}$$

the solution to $\mathbf{w}$ can be derived. More detailed, with a log operation and Gaussian assumptions, and by maximizing $J$ against $\mathbf{w}$ we have

$$\begin{aligned}
\nabla J &= \nabla \left( \left(\mathbf{y}-\mathbf{F}^T\mathbf{w}\right)^T \left(\mathbf{y}-\mathbf{F}^T\mathbf{w}\right) + \gamma \mathbf{w}^T\mathbf{w} \right) \\
&= \nabla \left( \mathbf{y}^T\mathbf{y} + \mathbf{w}^T\mathbf{F}\mathbf{F}^T\mathbf{w} - \mathbf{y}^T\mathbf{F}^T\mathbf{w} - \mathbf{w}^T\mathbf{F}\mathbf{y} + \gamma\mathbf{w}^T\mathbf{w} \right) \\
&= 2\mathbf{F}\mathbf{F}^T\mathbf{w} - 2\mathbf{F}\mathbf{y} + 2\gamma\mathbf{w} = 0
\end{aligned} \tag{23}$$

which directly gives the estimation of $\mathbf{w}$ as

$$\mu_\mathbf{w} = \left(\mathbf{F}\mathbf{F}^T + \gamma\mathbf{I}\right)^{-1} \mathbf{F}\mathbf{y} \tag{24}$$

$$\tag{25}$$

## B Derivation of $\Sigma_\mathbf{w}$

This part gets the variance of linear weights $\mathbf{w}$. For simplicity, re-write $\mu_\mathbf{w}$ as

$$\mu_\mathbf{w} = \left(\mathbf{F}\mathbf{F}^T + \gamma\mathbf{I}\right)^{-1} \mathbf{F}\mathbf{y} = \mathbf{M}\mathbf{y} \tag{26}$$

where $\mathbf{M} = \left(\mathbf{F}\mathbf{F}^T + \gamma\mathbf{I}\right)^{-1} \mathbf{F}$, $\mathbf{M} \in \mathbb{R}^{F\times N}$, $\mathbf{y} \in \mathbb{R}^{N\times 1}$, then the variance of $\mathbf{w}$ is expressed as

$$\begin{aligned}
\Sigma_\mathbf{w} &= \mathbf{M}\Sigma_\mathbf{y}\mathbf{M}^T \\
&= \Sigma_\mathbf{y}\mathbf{M}\mathbf{M}^T \\
&= \left(\mathbf{F}\mathbf{F}^T + \gamma\mathbf{I}\right)^{-1} \mathbf{F}\mathbf{F}^T \left(\left(\mathbf{F}\mathbf{F}^T + \gamma\mathbf{I}\right)^{-1}\right)^T \\
&= \left(\mathbf{F}\mathbf{F}^T + \gamma\mathbf{I}\right)^{-1} \mathbf{F}\mathbf{F}^T \left(\mathbf{F}\mathbf{F}^T + \gamma\mathbf{I}\right)^{-1} \\
&= \left(\mathbf{F}\mathbf{F}^T + \gamma\mathbf{I}\right)^{-1} \left(\mathbf{F}\mathbf{F}^T + \gamma\mathbf{I} - \gamma\mathbf{I}\right) \left(\mathbf{F}\mathbf{F}^T + \gamma\mathbf{I}\right)^{-1} \\
&= \left(\mathbf{F}\mathbf{F}^T + \gamma\mathbf{I}\right)^{-1} \left(\mathbf{F}\mathbf{F}^T + \gamma\mathbf{I}\right) \left(\mathbf{F}\mathbf{F}^T + \gamma\mathbf{I}\right)^{-1} - \gamma\mathbf{I}\left(\mathbf{F}\mathbf{F}^T + \gamma\mathbf{I}\right)^{-1}\left(\mathbf{F}\mathbf{F}^T + \gamma\mathbf{I}\right)^{-1} \\
&= \left(\mathbf{F}\mathbf{F}^T + \gamma\mathbf{I}\right)^{-1} - \gamma\left(\mathbf{F}\mathbf{F}^T + \gamma\mathbf{I}\right)^{-1}\left(\mathbf{F}\mathbf{F}^T + \gamma\mathbf{I}\right)^{-1}
\end{aligned} \tag{27}$$
$$\tag{28}$$

wherein $\Sigma_\mathbf{y} = \epsilon\mathbf{I}$ is a scaled identity matrix and for simplicity we can assume $\epsilon = 1$ without losing generality. Now given a new data point $\mathbf{f}$ for inference, we can get uncertainty as $\sigma = \mathbf{f}^T\Sigma_\mathbf{w}\mathbf{f}$. Name the eigen value of matrix $\left(\mathbf{F}\mathbf{F}^T + \gamma\mathbf{I}\right)$ as $\lambda$, we get the eigen value of matrix $\left(\mathbf{F}\mathbf{F}^T + \gamma\mathbf{I}\right)^{-1}$ as $\lambda^{-1}$.

As analyzed in the previous session, it is approximately $\lambda^{-1} \approx \frac{1}{N+\gamma}$ which is a infinitesimal associated with data amount $N$. Thus, $\lambda^{-2}$ can be negligible compared to $\lambda^{-1}$. Then we have uncertainty as

$$\begin{aligned}
\sigma &= \mathbf{f}^T \Sigma_{\mathbf{w}} \mathbf{f} \\
&= \mathbf{f}^T \left(\mathbf{F}\mathbf{F}^T + \gamma \mathbf{I}\right)^{-1} \mathbf{f} - \gamma \mathbf{f}^T \left(\mathbf{F}\mathbf{F}^T + \gamma \mathbf{I}\right)^{-1} \left(\mathbf{F}\mathbf{F}^T + \gamma \mathbf{I}\right)^{-1} \mathbf{f} \\
&= \left(\lambda^{-1} + \gamma \lambda^{-2}\right) \mathbf{f}^T \mathbf{f} \\
&\approx \lambda^{-1} \mathbf{f}^T \mathbf{f} \\
&= \mathbf{f}^T \left(\lambda^{-1} \mathbf{f}\right) \\
&= \mathbf{f}^T \left(\mathbf{F}\mathbf{F}^T + \gamma \mathbf{I}\right)^{-1} \mathbf{f}
\end{aligned} \tag{29}$$

This directly indicates that we can reduce $\Sigma_{\mathbf{w}}$ into

$$\Sigma_{\mathbf{w}} \approx \left(\mathbf{F}\mathbf{F}^T + \gamma \mathbf{I}\right)^{-1} \tag{30}$$

in our scenario because our uncertainty is $\sigma = \mathbf{f}^T \Sigma_{\mathbf{w}} \mathbf{f}$.

## C  Derivation of approximately $\sigma \propto \sqrt{F}$

$$\sigma^2 = \mathbf{f}^T \mathbf{A}^{-1} \mathbf{f} \tag{31}$$

$$= \mathbf{f}^T \mathbf{U}^T \Lambda \mathbf{U} \mathbf{f} \tag{32}$$

$$= \tilde{\mathbf{f}}^T \begin{bmatrix} \frac{1}{N+1} & 0 & 0 & 0 \\ 0 & 1 & 0 & 0 \\ 0 & 0 & \ddots & 0 \\ 0 & 0 & 0 & 1 \end{bmatrix} \tilde{\mathbf{f}} \tag{33}$$

$$= \begin{bmatrix} \tilde{\mathbf{f}}_1, \tilde{\mathbf{f}}_2, \cdots, \tilde{\mathbf{f}}_F \end{bmatrix} \begin{bmatrix} \frac{1}{N+1} & 0 & 0 & 0 \\ 0 & 1 & 0 & 0 \\ 0 & 0 & \ddots & 0 \\ 0 & 0 & 0 & 1 \end{bmatrix} \begin{bmatrix} \tilde{\mathbf{f}}_1 \\ \tilde{\mathbf{f}}_2 \\ \cdots \\ \tilde{\mathbf{f}}_F \end{bmatrix} \tag{34}$$

$$\approx F \|\tilde{\mathbf{f}}\|^2 \propto F \tag{35}$$

which is

$$\sigma \propto \sqrt{F} \tag{36}$$

Here we ignored the scaling factor $\epsilon$ or $\gamma$ as in Eq. 11 without losing generality but to make the derivation more succinct, and

$$\Lambda = \begin{bmatrix} \frac{1}{N+1} & 0 & 0 & 0 \\ 0 & 1 & 0 & 0 \\ 0 & 0 & \ddots & 0 \\ 0 & 0 & 0 & 1 \end{bmatrix} \tag{37}$$

$$\tilde{\mathbf{f}} = \mathbf{U} \mathbf{f} \tag{38}$$

where $\tilde{\mathbf{f}} = \begin{bmatrix} \tilde{\mathbf{f}}_1 \cdots \tilde{\mathbf{f}}_F \end{bmatrix}^T$, $\tilde{\mathbf{f}} \in \mathbb{R}^{F \times 1}$.